\documentclass[conference]{IEEEtran}

\IEEEoverridecommandlockouts
\usepackage{cite}
\usepackage{amsmath,amssymb,amsfonts}
\usepackage{graphicx}
\usepackage{textcomp}
\usepackage{xcolor}
\def\BibTeX{{\rm B\kern-.05em{\sc i\kern-.025em b}\kern-.08em
    T\kern-.1667em\lower.7ex\hbox{E}\kern-.125emX}}

\usepackage[utf8]{inputenc} % allow utf-8 input
\usepackage[T1]{fontenc}    % use 8-bit T1 fonts
\usepackage{hyperref}       % hyperlinks
\usepackage{url}            % simple URL typesetting
\usepackage{booktabs}       % professional-quality tables
\usepackage{amsfonts}       % blackboard math symbols
\usepackage{nicefrac}       % compact symbols for 1/2, etc.
\usepackage{microtype}      % microtypography

\usepackage{graphicx}

\usepackage{amsmath}
\usepackage{bbm}
\usepackage[colorinlistoftodos]{todonotes}
\usepackage{graphics} % for pdf, bitmapped graphics files
\usepackage{comment}
\usepackage{epsfig} % for postscript graphics files
\usepackage{amsmath} % assumes amsmath package installed
\usepackage{amssymb}  % assumes amsmath package installed
\usepackage{multicol}
\usepackage{multirow}
\usepackage{color}
\usepackage{mathrsfs}
\usepackage{xparse}
\usepackage{bm}
\usepackage{algorithm}
\usepackage{algorithmicx}
\usepackage{algpseudocode}
\usepackage[font=small,labelfont=bf,tableposition=top]{caption}
\usepackage[font=footnotesize]{subcaption}

\usepackage{mathtools}

\usepackage[colorinlistoftodos]{todonotes}

\usepackage{etoolbox}
\newtoggle{comments}
\settoggle{comments}{true} %set true to show all comments, set false to hide all comments
\iftoggle{comments}{
  \newcommand {\alberto}[1]{{\color{orange}{~Alberto: #1}\normalfont}}
  \newcommand {\bjin}[1]{{\color{blue}{~Baihong: #1}\normalfont}}
  \newcommand {\dan}[1]{{\color{SeaGreen}{~Dan: #1}\normalfont}}
  \newcommand {\yuxin}[1]{{\color{violet}{~Yuxin: #1}\normalfont}}
  \newcommand {\poolla}[1]{{\color{green}{~Kameshwar: #1}\normalfont}}
}{
  \newcommand {\alberto}[1]{{}}
  \newcommand {\bjin}[1]{{}}
  \newcommand {\dan}[1]{{}}
  \newcommand {\yuxin}[1]{{}}
  \newcommand {\poolla}[1]{{}}
}

\usepackage{acronym}
\acrodef{FTL}{Follow The Leader}
\acrodef{FTRL}{Follow The Regularized Leader}
\acrodef{SVM}{Support Vector Machine}
\acrodef{AE}{AutoEncoder}
\acrodef{RNN}{Recurrent Neural Network}
\acrodef{RUL}{Remaining Useful Life}
\acrodef{OC-SVM}{One-class Support Vector Machine}
\acrodef{CPS}{Cyber-Physical System}
\acrodef{DE}{Differential Evolution}
\acrodef{RBF}{Radial Basis Function}
\acrodef{EoL}{End of Life}
\acrodef{FDD}{Fault Detection and Diagnosis}

\begin{document}

\title{\LARGE \bf
A One-Class Support Vector Machine Calibration Method for Time Series Change Point Detection 
%\thanks{This work is supported in part by the National Research Foundation of Singapore through a grant to the Berkeley Education Alliance for Research in Singapore (BEARS) for the Singapore-Berkeley Building Efficiency and Sustainability in the Tropics (SinBerBEST) program, and by the National Science Foundation under Grant No.~1645964.}
}

\author{Baihong Jin$^1$~~Yuxin Chen$^2$~~Dan Li$^3$~~{Kameshwar Poolla}$^{1}$~~{Alberto Sangiovanni-Vincentelli}$^{1}$\\% <-this % stops a space
$^{1}$Department of EECS, University of California, Berkeley\\
{\tt\small \texttt{\{bjin,poolla,alberto\}@eecs.berkeley.edu}}\\
$^{2}$California Institute of Technology\\
{\tt\small \texttt{chenyux@caltech.edu}}\\
$^{3}$Institute of Data Science, National University of Singapore\\
{\tt\small \texttt{idsld@nus.edu.sg}}
}

\IEEEoverridecommandlockouts
\IEEEpubid{\makebox[\columnwidth]{978-1-5386-8357-6/19/\$31.00~\copyright2019 IEEE \hfill} \hspace{\columnsep}\makebox[\columnwidth]{ }}
\maketitle
\IEEEpubidadjcol

\begin{abstract}
It is important to identify the change point of a system's health status, which usually signifies an incipient fault under development. The One-Class Support Vector Machine (OC-SVM) is a popular machine learning model for anomaly detection and hence could be used for identifying change points; however, it is sometimes difficult to obtain a good OC-SVM model that can be used on sensor measurement time series to identify the change points in system health status. In this paper, we propose a novel approach for calibrating OC-SVM models. The approach uses a heuristic search method to find a good set of input data and hyperparameters that yield a well-performing model. Our results on the C-MAPSS dataset demonstrate that OC-SVM can also achieve satisfactory accuracy in detecting change point in time series with fewer training data, compared to state-of-the-art deep learning approaches. In our case study, the OC-SVM calibrated by the proposed model is shown to be useful especially in scenarios with limited amount of training data.
\end{abstract}


\section{INTRODUCTION}

\begin{figure}[t]
    \centering
    \includegraphics[width=0.7\columnwidth]{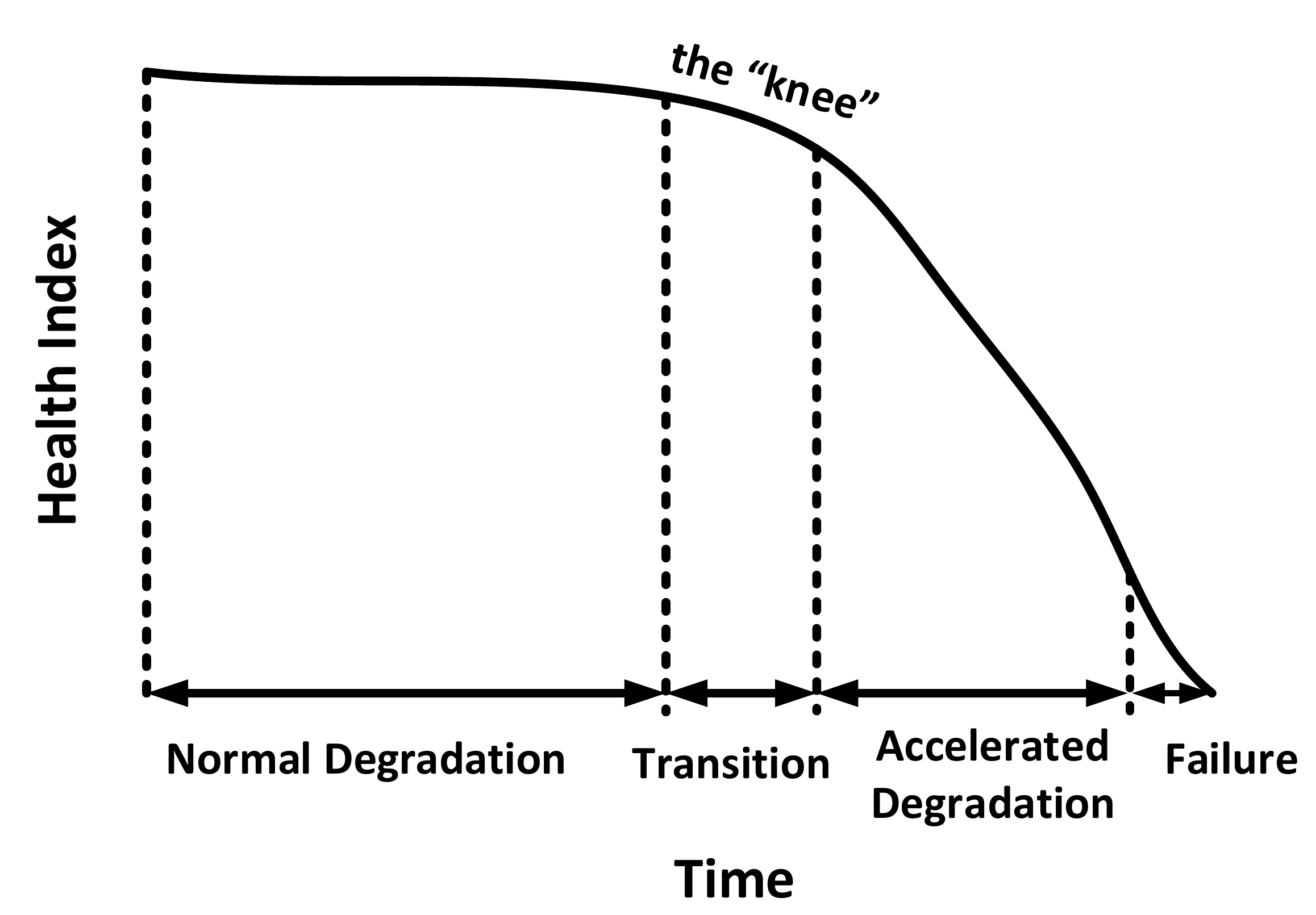}
    \caption{The four stages in a typical degradation process~\cite{ramasso2014review}.}
    \label{fig:four-stages}
\end{figure}

%Outliers are instances in a dataset that obviously deviate from the norm~\cite{goldstein2016comparative}. Identifying possible outliers are of great importance and interest in many real-world applications, including fault detection, fraud detection, and intrusion detection.
In system health monitoring, it is important to identify the change point, which usually signifies an incipient fault under development~\cite{d2011fuzzy}. Change point detection aims to find out when a system starts to shift away from its normal health condition into a faulty state. From a preventive maintenance perspective, it indicates that a maintenance action should be taken soon to intervene the degradation process in order to avoid further damage~\cite{Jin1906:Detecting}. Although system degradation is a gradual and complicated process, its development can in general be segmented into four discrete stages~\cite{ramasso2014review}: 1) normal degradation, 2) transition region, 3) accelerated degradation, and 4) failure; see Fig.~\ref{fig:four-stages} for a detailed illustration. We are interested in locating the transition region, a.k.a. the ``knee'' of the trajectory of the degrading health index.  

%suffer from two weaknesses when applied to complex systems: 1) they rely on a prior parametric model of the time series data, and 2) they often utilize simple features extracted from the input data such as the mean, variance and spectrum. Such detection methods 
In practice, change point detection is often a challenging task, especially in \ac{CPS} applications. As pointed out by authors of~\cite{lee2018time}, conventional change point detection models are often based on strong prior assumptions about the generative process that produces the observed data; however, the assumptions may not hold true for actual systems, which results in unsatisfactory performance of such methods. As another approach to tackle this challenge, various learning-based approaches have also been employed to distinguish different states in time series data. These approaches can be generally classified into \textit{supervised} and \textit{unsupervised} learning methods. In supervised learning, each data point is associated with a label that tells the category the point belongs to. Based on the supervisory information, a supervised learning algorithm infers a decision function that can be used to assign labels to data points; ideally, the learned decision function can generalize well to unseen new data points. In the \ac{FDD} literature, supervised methods have been shown to be effective at distinguishing healthy and faulty states, when labels for training data are available. In fact, in real applications it is very difficult to obtain labels that accurately indicate the change point locations of training data, making it a challenge to apply supervised methods for change point detection. Even if we have detected the existence of faults, we do not know when the fault starts to develop. Unsupervised approaches, e.g.~One-Class Support Vector Machine (OC-SVM)~\cite{scholkopf2001estimating} and autoencoders~\cite{sakurada2014anomaly}, are more suitable in such scenarios where there are not enough labeled data for differentiating a system's health status. Still, in OC-SVM and autoencoder approaches, we need to provide a training dataset that comprise of mostly fault-free data. In addition, the choice of hyperparameter may also greatly impact the model's performance. Without labeled data for cross-validation, it is a challenging task to select the input data and the hyperparameters.

% using only unlabeled data
In this paper, we address these challenges by proposing a heuristic approach for \textit{calibrating} OC-SVM models.
During the calibration, besides the hyperparameters, we also controlled how much training data is used for training the OC-SVM model. The calibrated OC-SVM model is used for detecting the change points in timeseries data. In particular, we study scenarios where: 1) the system can be assumed to be healthy from the start of until a point where faults start to develop, and ends up in a fault state, 2) the degradation can be viewed as an {approximately monotonic} process. We believe that such setup is representative of many real-world degradation processes. To validate our proposed approach, we conducted a case study on the C-MAPSS dataset~\cite{saxena2008c}, and demonstrated strong empirical performance in detecting change points. 

The remainder of this paper is organized as follows. In Sec.~\ref{sec:background}, we give necessary background on OC-SVM. Next in Sec.~\ref{sec:approach}, we discuss on our proposed search framework in details. A brief description about the C-MAPSS flight engine dataset is provided in Sec.~\ref{sec:cmapss}, and experimental results is presented in Sec.~\ref{sec:experiment}. We conclude the paper in Sec.~\ref{sec:conclusion} and discuss future directions. 
\section{One-Class Support Vector Machine}\label{sec:background}
A one-class SVM (OC-SVM)~\cite{scholkopf2001estimating} model denotes an unsupervised machine learning model that learns a decision function for estimating the support of a dataset. This property makes it applicable to outlier detection. In literature, it has been applied to areas such as fault detection, intrusion detection, and forgery signature detection~\cite{rosso2016classification}. An OC-SVM model is trained on data mostly from one class (usually the ``normal'' class). The trained model can be used to classify new data as similar or different to the training set. This is useful for outlier detection because there are usually very few examples for outliers (anomalies or faults), which makes it difficult to train a two-class classifier to distinguish them. 

To obtain more versatile decision boundaries with OC-SVM, kernel functions are often used to map the original feature space into higher-dimensional spaces. Let $\Phi:\mathcal{X}\rightarrow\mathcal{Z}$ be a mapping from the original data space to a higher-dimensional feature space. To train an OC-SVM model, we solve the following quadratic program.
\begin{align}
    &\min_{\bm{w},\xi_i,b} \frac{1}{2}\Vert\bm{w}\Vert_2^2 + \frac{1}{\nu m}\sum_{i=1}^m\xi_i-b\\
    &\text{s.t.}~\forall i=1,2,\ldots,N\notag\\
    &\quad\quad\nu\in(0,1], \xi_i\geq 0,\notag\\
    &\quad\quad\bm{w}\cdot\Phi(z_i)\geq b-\xi_i.\notag
\end{align}
Where $\xi_i$ is a slack variable for data point $i$. $\nu$ takes a value between 0 and 1; it upper bounds the fraction of training errors and lower bounds the fraction of support vectors. As a result, a non-zero $\nu$ will allow an OC-SVM model to exclude some of the training data as outliers from the normal class. We use the following decision function, to decide whether or not an input $x$ belongs to the normal class.
\begin{align}
    f(z) = \text{sign}\left(\bm{w}\cdot\Phi(x)-b\right)
    &=\begin{cases}
    +1~\text{normal},\\
    -1~\text{outlier}.
    \end{cases}
    %&= \sum_{i=1}^N \alpha_i K(x,x_i)
\end{align}

In this study, we choose \ac{RBF} kernels in our OC-SVM formulation because they do not assume any parametric form of the data distribution, and thus they are suitable for capturing data of complex shape. A \ac{RBF} kernel has a form $K(z_i,z_j) = \exp\left(-\gamma\Vert {z_i-z_j}\Vert^2\right)$, where $\gamma$ controls the width of the \ac{RBF} kernel used for training. The larger $\gamma$ is, the smaller the width of the kernel. If gamma is too large, the model will overfit the data and will not capture the shape of the data distribution.

\section{Heuristic OC-SVM Calibration Approach}\label{sec:approach}

%\dan{section title better to be more specific, use"Heuristic Model Validation"?}

Both the input training data and the choice of OC-SVM hyperparameters can influence the prediction performance of OC-SVM models. The choice of the two hyperparameters $\nu$ and $\gamma$ can greatly affect the prediction performance of the model. In fact, these hyperparameters are coupled, because  $\nu$ regulates the percentage of outliers in the training data. If we choose a small $\nu$, then we must ensure that $1-\nu$ of the training data are from the positive class. We are facing a dilemma here: to obtain a good OC-SVM model for change point detection, we must ensure that the training data contains few outliers, which requires us to know the location of the change points \textit{a priori}. To resolve the dilemma, we resort to a heuristic scheme to search for a suitable assignment of input data and hyperparameters as our hypothesis, such that the predictive outcome matches the hypothesis itself. Since the training of OC-SVM is achieved by solving a quadratic optimization problem, it can be done efficiently with state-of-the-art optimization tools. The fast training time of OC-SVM enables us to use a heuristic optimization approach to search for an optimal configuration, which will be introduced next.

Let us index the $m$ system instances under study by $1,2,\ldots,m$. By assumption, the systems all degrade in an approximately monotonic fashion. We designate their respective change points by a vector $\bm{\rho}=(\rho_1,\rho_2,\ldots,\rho_m)$, where each $\rho_i\in(0,1)$ is a number representing the portion of life span when the $i$th system hits its change point. For example, $\rho_i=0.7$ means that engine $i$ is healthy in the first $70\%$ of its life span before reaching the change point. 
By fixing $\nu$ to a small positive value, we look for an assignment of $\bm{\rho}$ where the input data are mostly healthy (from the positive class). The decision variable $(\gamma,\bm{\rho})$ now consists of $m+1$ dimensions. A sensible hypothesis of $(\gamma,\bm{\rho})$ will yield an OC-SVM model whose prediction will agree with the hypothesis. We use a \ac{DE}-based heuristic search scheme to find  $(\gamma,\bm{\rho})$. For guiding the heuristic search, a loss function is needed to evaluate the \textit{goodness of fit} of each hypothesized hyperparameters. Analogous to hypothesis testing in statistics, if a hypothesis $(\gamma,\bm{\rho})$ does not yield a prediction that is \textit{consistent} with itself, we can claim that the hypothesis is probably not true.

\subsection{Loss Function for Examining the ``Goodness of Fit'' of a Hypothesis}
In our scenario, each hypothesized $\rho_i$ signifies the position of change point (as a percentage value) of training system instance $i$. Let the actual position (in cycles) of the change point be $c_i$. 
%Cycles are not defined. Please do
In other words, the first $c_i$ cycles can be assumed healthy (labeled $1$), while the rest are assumed unhealthy (labeled $0$). For each hypothesis $(\gamma,\bm{\rho})$, we will check how consistent it is with the learned OC-SVM classifier and its associated classification results by using the loss function as described below.
\begin{align}\label{eqn:loss}
    L(\gamma,\bm{\rho}) = \sum_{i=1}^{m} \text{log\_loss}\left(\bm{y}^{(i)},\bm{z}^{(i)}\right),
\end{align}
where $\bm{y}^{(i)}$ is the hypothesized label assignment with the first $c_i$ elements being $1$'s, and otherwise $0$'s. $\bm{z}^{(i)}$ is the predicted label assignment from the trained classifier.

When a trained classifier is applied to an unseen system instance, its change point can be derived by finding the point in life that gives the lowest cross-entropy loss (i.e.,~log loss).

\subsection{Differential Evolution for Hyperparameter Optimization}

%\acf{DE} is a heuristics for solving global optimization problems and belongs to the general category of evolutionary algorithms. 

\ac{DE} is an evolutionary algorithm that is commonly used for solving global optimization problems.
The \ac{DE} optimizes a problem by maintaining a population of candidate solutions and creating new candidate solutions by combining existing ones according to some simple formulae. Candidate solutions that receive better scores are kept and will be used for generating new candidate solutions during the next generation. The optimization problem is thus treated as a black box, and the \ac{DE} procedure does need gradient information to optimize it. With the loss function described above, we can use \ac{DE} to search for a hypothesis whose prediction result is consistent with the hypothesized hyperparameter assignment itself. It is worthy to note that there does not need to be an ``accurate'' optimum, because the change point actually is an interval instead of a single point. Our purpose is achieved, as long as the found change point falls within the transition region. 

In this work, we use \ac{DE}~\cite{storn1997differential,price2006differential} as our search framework. It is worthy to note that, besides \ac{DE}, other heuristic search algorithms such as Bayesian optimization~\cite{pelikan1999boa,snoek2012practical} and simulated annealing~\cite{kirkpatrick1983optimization} may also be applied here. It is part of our ongoing work to test out the effectiveness of these other search algorithms. 

% The pseudocode for the hyperparameter search procedure is described in Algo.~\ref{algo:DE}. 
% \begin{algorithm}[b]
% \caption{Differential Evolution for Hyperparameter Optimization}\label{algo:DE}
% \begin{algorithmic}
% \State Create initial population
% \While{termination condition not reached}
% \State evaluate goodness of fit
% \State mutate candidate solution
% \EndWhile
% \end{algorithmic}
% \end{algorithm}
\section{C-MAPSS DATASET}\label{sec:cmapss}

\subsection{Organization of the C-MAPSS Dataset}

The C-MAPSS dataset~\cite{saxena2008c} was originally released in the PHM-08~Challenge. The dataset was generated by a high-fidelity engine simulator. During the simulation for each engine, some fault was injected at a given time and persisted throughout the remaining flights. The degradation process over the lifespan of an engine is described as a multivariate timeseries consisting of a few hundred data points (cycles). Each data point is a 21-dimension vector corresponding to 21 sensor readings. Because little system-specific knowledge about the engines was given, useful information needs to be mined from data themselves, making the learning problem a data-driven task. Since each timeseries describes the run-to-failure process of an engine, we also know the \ac{RUL} of the engine at each time instant.  

\begin{figure}[b]
    \centering
    \includegraphics[width=0.8\columnwidth]{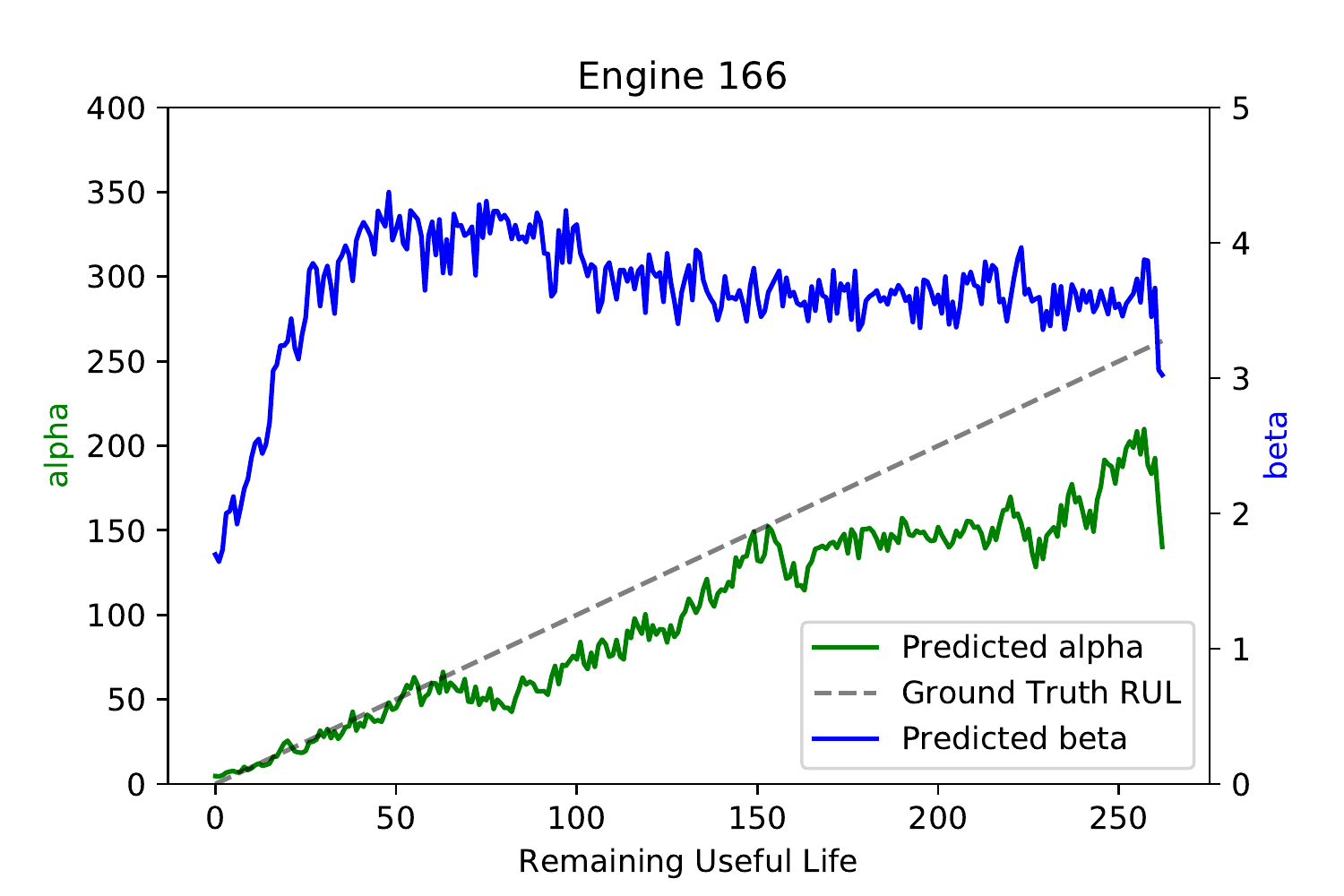}
    \caption{We implemented and trained a WTTE-RNN model in Keras, using all $249$ engines from \texttt{FD004}. Trends of predicted $\alpha$ and $\beta$ on engine instance 166.}\label{fig:individual-engine}
\end{figure}

\begin{figure*}[t]
    \centering
    \begin{subfigure}[b]{0.45\linewidth}
        \centering
        \includegraphics[width=\textwidth]{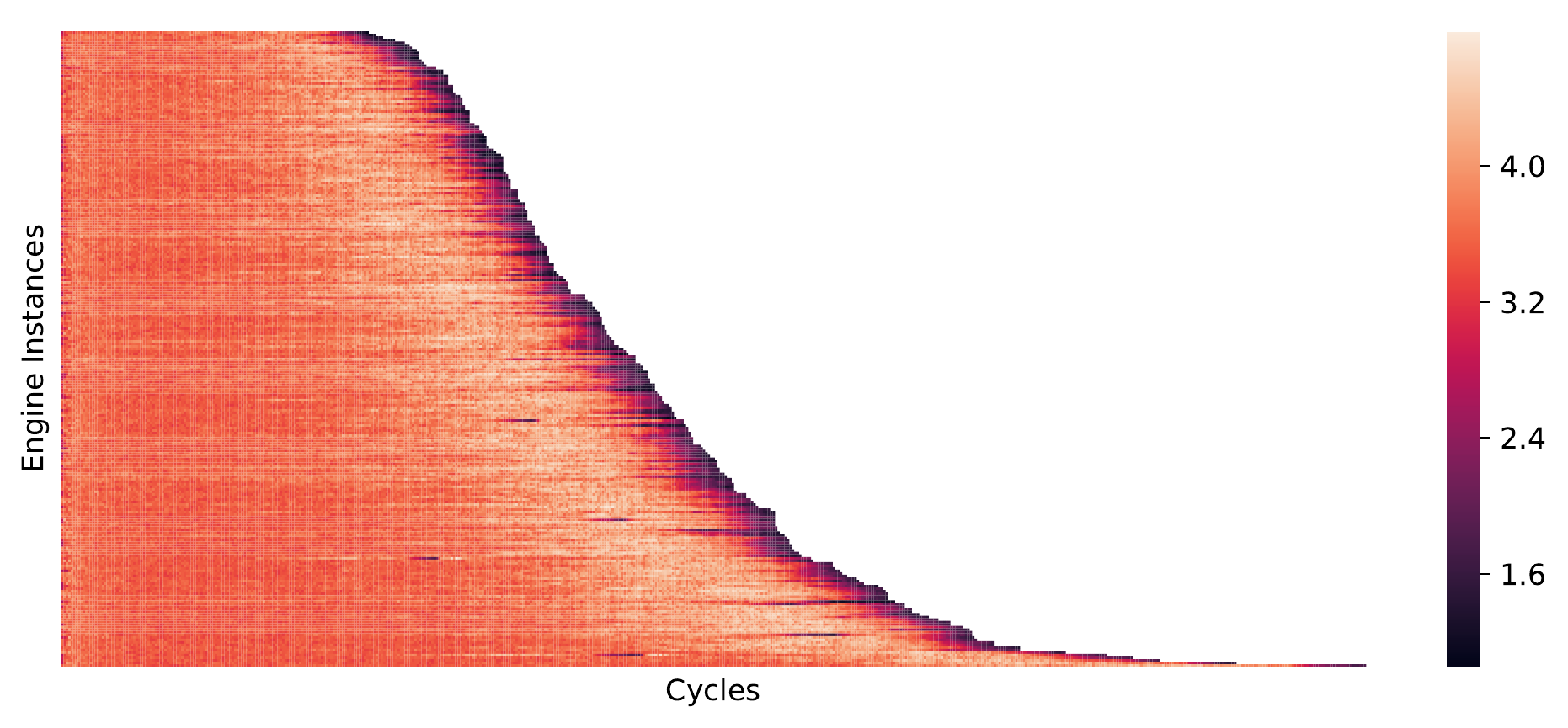}
        \caption{}\label{fig:beta-all-engines}
    \end{subfigure} 
    ~
    \begin{subfigure}[b]{0.45\linewidth}
        \centering
        \includegraphics[width=\textwidth]{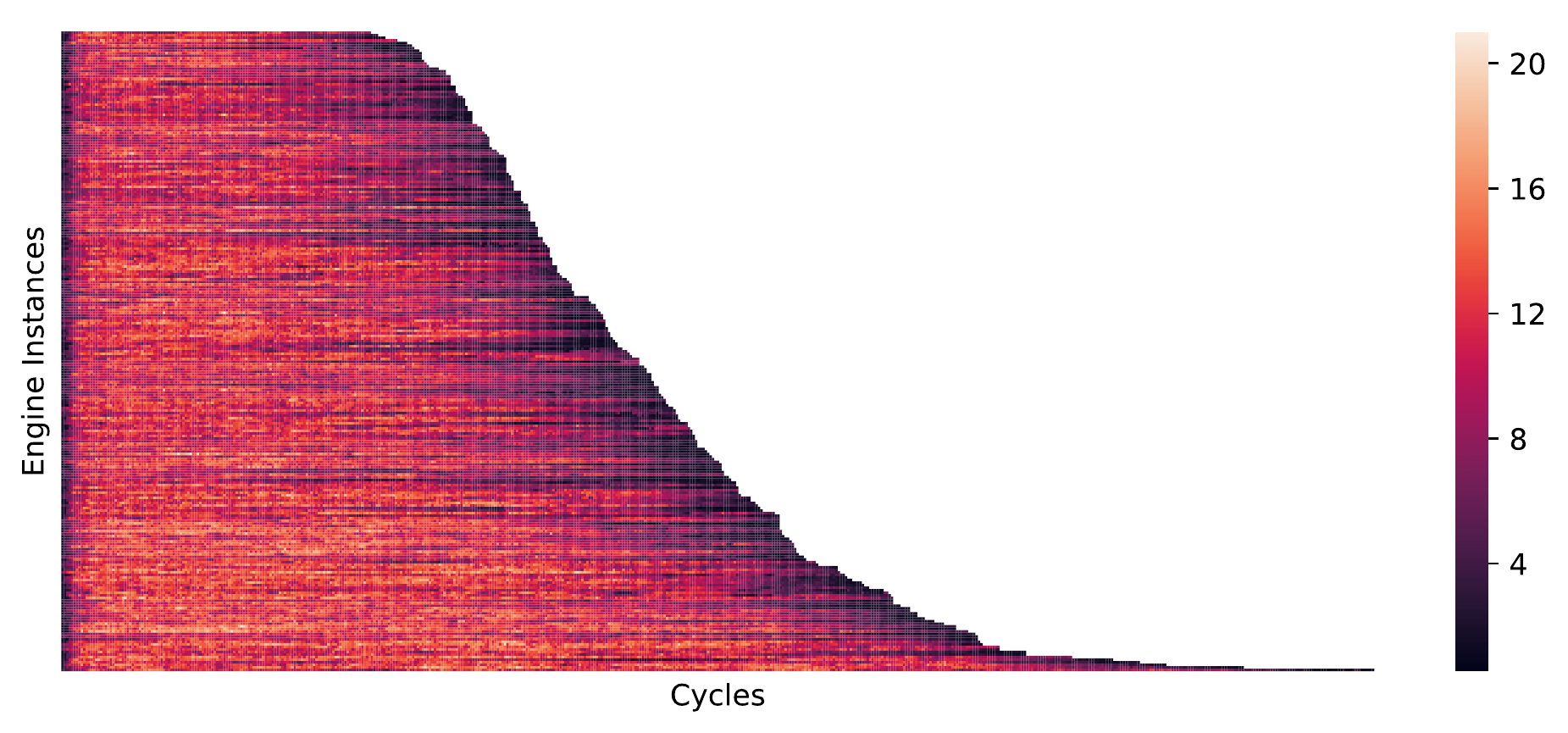}
        \caption{}\label{fig:beta-all-engines-few-data}
    \end{subfigure}
    \caption{(a) The predicted $\beta$ values on all 249 engine instances using our approach (b) The $\beta$ values predicted by another WTTE-RNN model that is trained using only 20 engine instances for comparison purposes. The $\beta$ values are signified by different colors.}\label{fig:WTTE-RNN-results}
\end{figure*}

The original goal of the PHM-08 Challenge was to stimulate the development of data-driven \ac{RUL} estimation methods. Since the release of the C-MAPSS dataset, many different \ac{RUL} estimation approaches are seen in literature. A comprehensive review of the published results can be found in~\cite{ramasso2014review}.
The dataset consists of four parts (\texttt{FD001}-\texttt{FD004}) that represent an increasing level of complexity. \texttt{FD001}-\texttt{FD003} can be seen as special cases of \texttt{FD004}. The \texttt{FD001} data represent the simplest scenario, where the engines are operating only at sea level, and are injected with only one type of fault. The \texttt{FD004} data involves two fault types and six flight conditions. The review paper~\cite{ramasso2014review} noticed that many publications used only \texttt{FD001}, the most basic scenario, for validating their algorithms; their actual performance on more realistic scenarios such as \texttt{FD004} is thus doubtful. Because of this, the \texttt{FD004} data are used to validate our approach in the experiment.

Given the ground truth \ac{RUL} number at each observed data point, the \ac{RUL} prediction problem can be formulated as a regression problem. Supervised learning methods, such as neural networks, can be used to establish a mapping between feature vectors and \ac{RUL}. Due to their ability to capture complex temporal relationship among data, a number of papers~\cite{heimes2008recurrent,zheng2017long} employed variants of \ac{RNN} to model the functional mapping and reported to yield satisfactory results.  

\subsection{Change Point Detection with WTTE-RNN}

%As previously mentioned, it is important to detect changes in system health status. 
An accurate \ac{RUL} regression model, however, does not directly serve our purpose of change point identification. Recently, Martinsson proposed a novel \ac{RNN} architecture called WTTE-RNN\footnote{The name WTTE-RNN stands for ``Weibull Time To Event Recurrent Neural Network''.}~\cite{martinsson:Thesis:2016} that provided additional insight into the degradation process. Not only can WTTE-RNN provide uncertainty estimation about the predicted \ac{RUL}, but it also indicates potential changes in the health status. Since the C-MAPSS dataset itself does not provide ground truth information about the change point location for each engine, so we need another approach for validating our proposed approach. In this study, we used the segmentation results given by the WTTE-RNN as the ground truth to benchmark our proposed OC-SVM approach.

Here we brief describe the WTTE-RNN approach. WTTE-RNN is a \ac{RNN} architecture for \textit{time-to-event} prediction. As with other \ac{RNN} architectures, a WTTE-RNN model makes prediction at each time step based on previous input sequences. Specifically, at each time step $t$, the model predicts two numbers $\alpha_t$ and $\beta_t$, the two parameters of a Weibull distribution that characterize the \ac{RUL} at time $t$. The predicted $\alpha_t$ parameter represents the expected \ac{RUL} value at time step $t$, and $\beta_t$ quantifies the associated uncertainty of the \ac{RUL} prediction. By feeding the observed timeseries data into the trained WTTE-RNN network, we will obtain sequences of predicted $\alpha$ and $\beta$ values, which reflect the updating prediction on the system health status. 

To illustrate the performance of WTTE-RNN on the C-MAPSS dataset, we implemented a WTTE-RNN model in Keras, and trained the network using all $249$ engines from \texttt{FD004}. As an example, Fig.~\ref{fig:individual-engine} plots the produced $\alpha$ and $\beta$ evaluations for engine 166. It can be seen that the produced $\alpha_t$ is able to predict the \ac{RUL} with good accuracy, especially when the engine is close to its \ac{EoL}. In addition, the $\beta$ values are large when the engine is healthy, and decreases rapidly when the engine moves towards the \ac{EoL}. Intuitively, it is usually difficult to predict \ac{RUL} accurately when an engine is healthy, resulting in relatively high $\beta$ values, and easier to predict \ac{RUL} when an engine moves towards its \ac{EoL}. In particular, we can also observe a short period of increase in $\beta$ value before it rapidly drops, which indicates higher uncertainties values in this region. We show the predicted $\beta$ value on all 249 engines as a heatmap in Fig.~\ref{fig:beta-all-engines}. It can be seen that the phenomenon mentioned above appears in almost all the engines. Martinsson conjectured~\cite{martinsson:Thesis:2016} that the elevated uncertainty values are due to the transition from the healthy state to a fast degradation phase, i.e.~the ``knee'' in Fig.~\ref{fig:four-stages}. More details about the WTTE-RNN approach are beyond the scope of this paper, and we refer interested readers to~\cite{martinsson:Thesis:2016} for a more in-depth discussion. 

Although WTTE-RNN has demonstrated promising results, the model when trained with data from only 20 engines fails to give satisfactory results. The resulting $\beta$ predictions, as displayed in Fig.~\ref{fig:beta-all-engines-few-data}, do not  clearly indicate the change points. This suggests that the performance of this approach might be affected when training data are limited. It will be shown shortly, our approach in comparison requires much fewer data, making it advantageous when the available amount of data is limited.
\section{Experimental Results}\label{sec:experiment}

%\yuxin{Do we have empirical results for this? It will be interesting to see a plot on performance vs data tradeoff, i.e., if we are only allowed to use the same amount of data as the OC-SVM for the WTTE-RNN approach, how bad will the performance be?} \bjin{We do have results, but it may take some space and effort to describe and explain them. Shall we save them for our later full paper submission?} \yuxin{sure. Works for me!}

We randomly selected $m=20$ out of a total of $249$ engines as the training set, and used the rest as the test set.  In our experiment, the population consisted of $105$ individuals, and was evolved for $10$ generations. We fixed $\nu=0.05$ during the heuristic search to reduce the complexity. To further reduce the search space, we assumed the first $50\%$ of the observed data for each training engine is healthy, i.e. $\forall i,~0.5\leq\rho_i\leq 1.0$. 
Because a good value for $\gamma$ could vary much (typical values between $10^{-2}$ and $10^2$), in our \ac{DE} procedure we search for $\log_{10}\gamma\in[-2,2]$ instead. We implement the algorithm in Python using the \texttt{OneClassSVM} module from \texttt{scikit-learn} for training OC-SVM classifiers, and the \texttt{differential\_evolution} module from \texttt{scipy} as the framework for implementing the \ac{DE} search procedure.

\begin{figure}[tb]
    \centering
    \includegraphics[width=0.9\columnwidth]{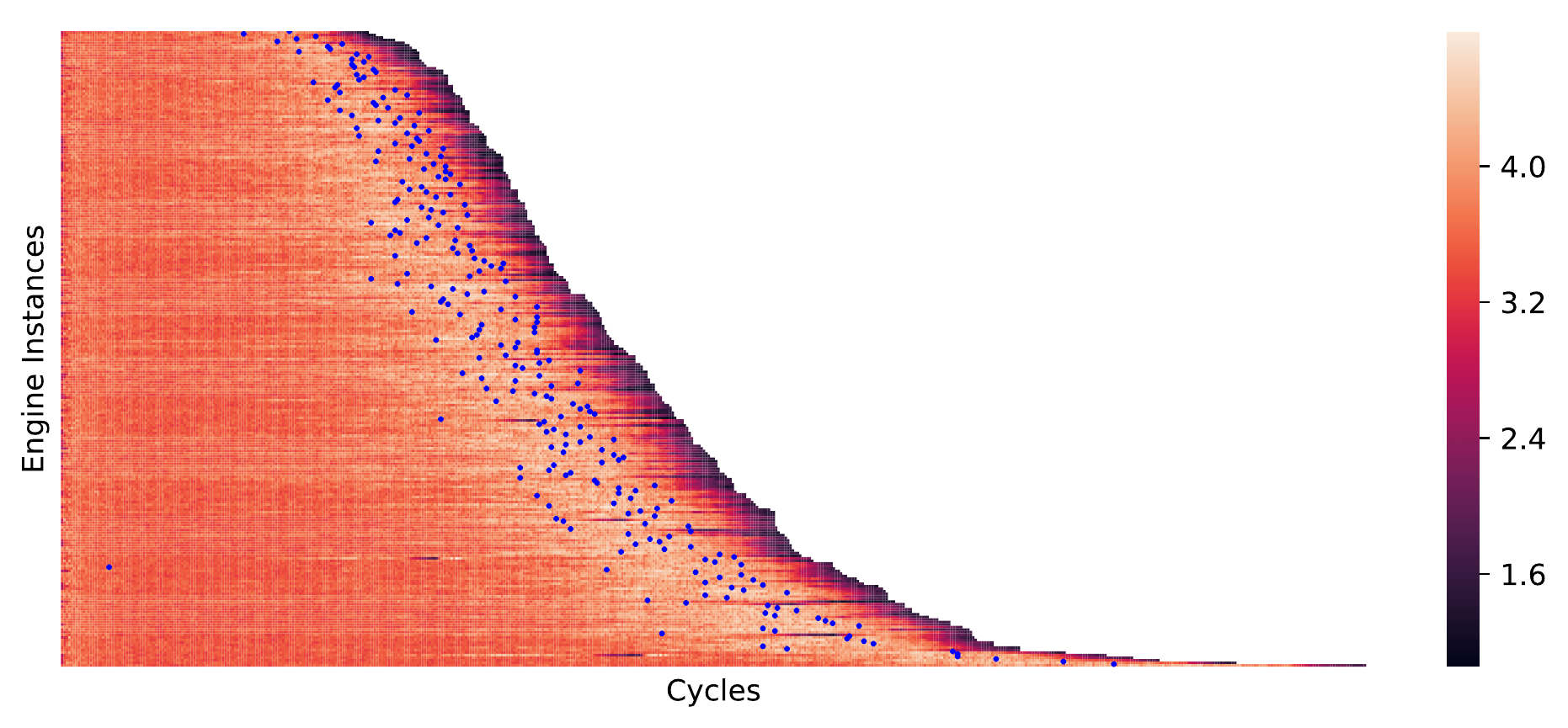}
    \caption{The change points (blue dots) identified by the OC-SVM approach for all 249 engine instances.}
    \label{fig:results}
\end{figure}

After evolving for 10 generations, the \ac{DE} search procedure returned the optimal OC-SVM model configuration with $\gamma=3.59$, as well as the change point locations for the 20 engines in the training set. To obtain the change point of the rest of the engines, we used the method mentioned in Sec.~\ref{sec:approach}. The identified change points by our approach are plotted as the blue points in Fig.~\ref{fig:results}. 
%\dan{Just a simple comment, the result is solid now, but could it be better if we can compare OC-SVM validated by the proposed method with OC-SVM with simple cross-validation?} 
As can be seen, for most of the engines, the change points identified by our algorithm match the transition region (white regions with high $\beta$ values) given by WTTE-RNN. Although the OC-SVM model lacks the ability to predict the \ac{RUL}, it still identifies the correct change point locations. As opposed to the WTTE-RNN that needs a large amount of data to train (249 engines was used in this example), our OC-SVM model was trained with data from only 20 engines. This shows that our proposed OC-SVM calibration approach, as a data-efficient method, can find its use in scenarios with limited available data.

%\bjin{Display prediction results from OC-SVM!!}

\section{Conclusions}\label{sec:conclusion}
OC-SVM is a popular unsupervised machine learning model for anomaly detection; however, there are practical concerns when applying this model for timeseries change point detection. The performance of the OC-SVM model depends highly on choice of both the training data and the hyperparameters, making it hard to train a good model in the absence of labeled data for cross validation. In this paper, we attempt to address this challenge by using a heuristic method to search for a suitable model that explains the data. Our experiments on the C-MAPSS dataset have shown promising results, which indicates the usefulness of our proposed approach as a data-efficient way to infer the location of change points in system degradation process.

\section*{Acknowledgment}
This work is supported in part by the National Research Foundation of Singapore through a grant to the Berkeley Education Alliance for Research in Singapore (BEARS) for the Singapore-Berkeley Building Efficiency and Sustainability in the Tropics (SinBerBEST) program, and by the National Science Foundation under Grant No.~1645964.

\medskip
\bibliographystyle{IEEEtran}
\bibliography{refs}

% Generated by IEEEtran.bst, version: 1.14 (2015/08/26)
\begin{thebibliography}{10}
\providecommand{\url}[1]{#1}
\csname url@samestyle\endcsname
\providecommand{\newblock}{\relax}
\providecommand{\bibinfo}[2]{#2}
\providecommand{\BIBentrySTDinterwordspacing}{\spaceskip=0pt\relax}
\providecommand{\BIBentryALTinterwordstretchfactor}{4}
\providecommand{\BIBentryALTinterwordspacing}{\spaceskip=\fontdimen2\font plus
\BIBentryALTinterwordstretchfactor\fontdimen3\font minus
  \fontdimen4\font\relax}
\providecommand{\BIBforeignlanguage}[2]{{%
\expandafter\ifx\csname l@#1\endcsname\relax
\typeout{** WARNING: IEEEtran.bst: No hyphenation pattern has been}%
\typeout{** loaded for the language `#1'. Using the pattern for}%
\typeout{** the default language instead.}%
\else
\language=\csname l@#1\endcsname
\fi
#2}}
\providecommand{\BIBdecl}{\relax}
\BIBdecl

\bibitem{d2011fuzzy}
M.~F. D'Angelo, R.~M. Palhares, R.~H. Takahashi, and R.~H. Loschi,
  ``Fuzzy/{B}ayesian change point detection approach to incipient fault
  detection,'' \emph{IET control theory \& applications}, vol.~5, no.~4, pp.
  539--551, 2011.

\bibitem{Jin1906:Detecting}
B.~Jin, D.~Li, S.~Srinivasan, S.-K. Ng, K.~Poolla, and
  A.~Sangiovanni-Vincentelli, ``Detecting and diagnosing incipient building
  faults using uncertainty information from deep neural networks,'' in
  \emph{2019 IEEE International Conference on Prognostics and Health Management
  (ICPHM) (PHM2019), submitted}, San Francisco, USA, Jun. 2019.

\bibitem{ramasso2014review}
E.~Ramasso and A.~Saxena, ``Review and analysis of algorithmic approaches
  developed for prognostics on {CMAPSS} dataset,'' in \emph{Annual Conference
  of the Prognostics and Health Management Society 2014.}, 2014.

\bibitem{lee2018time}
W.-H. Lee, J.~Ortiz, B.~Ko, and R.~Lee, ``Time series segmentation through
  automatic feature learning,'' \emph{arXiv preprint arXiv:1801.05394}, 2018.

\bibitem{scholkopf2001estimating}
B.~Sch{\"o}lkopf, J.~C. Platt, J.~Shawe-Taylor, A.~J. Smola, and R.~C.
  Williamson, ``Estimating the support of a high-dimensional distribution,''
  \emph{Neural computation}, vol.~13, no.~7, pp. 1443--1471, 2001.

\bibitem{sakurada2014anomaly}
M.~Sakurada and T.~Yairi, ``Anomaly detection using autoencoders with nonlinear
  dimensionality reduction,'' in \emph{Proceedings of the MLSDA 2014 2nd
  Workshop on Machine Learning for Sensory Data Analysis}.\hskip 1em plus 0.5em
  minus 0.4em\relax ACM, 2014, p.~4.

\bibitem{saxena2008c}
A.~Saxena and K.~Goebel, ``{C-MAPSS} data set,'' \emph{NASA Ames Prognostics
  Data Repository}, 2008.

\bibitem{rosso2016classification}
O.~A. Rosso, R.~Ospina, and A.~C. Frery, ``Classification and verification of
  handwritten signatures with time causal information theory quantifiers,''
  \emph{PloS one}, vol.~11, no.~12, p. e0166868, 2016.

\bibitem{storn1997differential}
R.~Storn and K.~Price, ``Differential evolution--a simple and efficient
  heuristic for global optimization over continuous spaces,'' \emph{Journal of
  global optimization}, vol.~11, no.~4, pp. 341--359, 1997.

\bibitem{price2006differential}
K.~Price, R.~M. Storn, and J.~A. Lampinen, \emph{Differential evolution: a
  practical approach to global optimization}.\hskip 1em plus 0.5em minus
  0.4em\relax Springer Science \& Business Media, 2006.

\bibitem{pelikan1999boa}
M.~Pelikan, D.~E. Goldberg, and E.~Cant{\'u}-Paz, ``Boa: The bayesian
  optimization algorithm,'' in \emph{Proceedings of the 1st Annual Conference
  on Genetic and Evolutionary Computation-Volume 1}.\hskip 1em plus 0.5em minus
  0.4em\relax Morgan Kaufmann Publishers Inc., 1999, pp. 525--532.

\bibitem{snoek2012practical}
J.~Snoek, H.~Larochelle, and R.~P. Adams, ``Practical bayesian optimization of
  machine learning algorithms,'' in \emph{Advances in neural information
  processing systems}, 2012, pp. 2951--2959.

\bibitem{kirkpatrick1983optimization}
S.~Kirkpatrick, C.~D. Gelatt, and M.~P. Vecchi, ``Optimization by simulated
  annealing,'' \emph{Science}, vol. 220, no. 4598, pp. 671--680, 1983.

\bibitem{heimes2008recurrent}
F.~O. Heimes, ``Recurrent neural networks for remaining useful life
  estimation,'' in \emph{Prognostics and Health Management, 2008. PHM 2008.
  International Conference on}.\hskip 1em plus 0.5em minus 0.4em\relax IEEE,
  2008, pp. 1--6.

\bibitem{zheng2017long}
S.~Zheng, K.~Ristovski, A.~Farahat, and C.~Gupta, ``Long short-term memory
  network for remaining useful life estimation,'' in \emph{2017 IEEE
  International Conference on Prognostics and Health Management (ICPHM)}.\hskip
  1em plus 0.5em minus 0.4em\relax IEEE, 2017, pp. 88--95.

\bibitem{martinsson:Thesis:2016}
E.~Martinsson, ``{WTTE-RNN : Weibull Time To Event Recurrent Neural Network},''
  Master's thesis, Chalmers University Of Technology, 2016.

\end{thebibliography}

\end{document}